\documentclass[10pt,journal, twocolumn]{IEEEtran}

\IEEEoverridecommandlockouts

\usepackage{cite}
\usepackage{amsmath,amssymb,amsfonts}
\usepackage{graphicx}
\usepackage{textcomp}
\usepackage[dvipsnames]{xcolor}

\usepackage{stackengine}

\usepackage{float}
\usepackage{bbm}
\usepackage{multirow}
\usepackage{fancyhdr, epsfig, epsf, amsthm, amsmath, amssymb, amsfonts,color,dsfont,epstopdf}
\usepackage{dblfloatfix} 
\usepackage[most]{tcolorbox}
\tcbuselibrary{breakable} 
\usepackage{enumitem}
\usepackage{url}
\usepackage{algorithm,algorithmic,xpatch}
\usepackage{booktabs}

\usepackage{subfigure}
\usepackage{indentfirst}
\usepackage{wrapfig,lipsum,booktabs}

\usepackage{cancel}
\usepackage{soul}
\usepackage{eucal}

\def\l{\left}
\def\r{\right}
\def\({\l(}
\def\){\r)}
\def\[{\l[}
\def\]{\r]}

\DeclareMathOperator*{\argmaxA}{argmax}   

\allowdisplaybreaks

\begin{document}

\title{Attention Based Communication and Control for Multi-UAV Path Planning}

\author{Hamid~Shiri, Hyowoon~Seo, $^\dagger$Jihong Park, and Mehdi~Bennis

\thanks{
H. Shiri, M. Bennis are with the Centre for Wireless Communications, University of Oulu, Oulu 90014, Finland (email:\{hamid.shiri, mehdi.bennis\}@oulu.fi).}
\thanks{H. Seo is with the Institute of New Media and Communications and Department of Electrical and Computer Engineering, Seoul National University (SNU), Seoul 08826, Korea (email:hyowoonseo@snu.ac.kr).}
\thanks{$^\dagger$J. Park is with the School of Information Technology, Deakin University, Geelong, VIC 3220, Australia (email: jihong.park@deakin.edu.au).
}
\vspace{-18pt}
}

\maketitle

\begin{abstract}
Inspired by the multi-head attention (MHA) mechanism in natural language processing, this letter proposes an iterative single-head attention (ISHA) mechanism for multi-UAV path planning. The ISHA mechanism is run by a communication helper collecting the state embeddings of UAVs and distributing an attention score vector to each UAV. The attention scores computed by ISHA identify how many interactions with other UAVs should be considered in each UAV's control decision-making. Simulation results corroborate that the ISHA-based communication and control framework achieves faster travel with lower inter-UAV collision risks than an MHA-aided baseline, particularly under limited communication resources.
\end{abstract}


\IEEEpeerreviewmaketitle

\section{Introduction}

 Unmanned aerial vehicle (UAV) is deemed to be one key component in Industry 4.0 and of non-terrestrial networks (NTNs) in beyond 5G/6G communication systems \cite{BinLi2019UAV5G, 8620938, shiri2020remote, MFGuav2019shiri}. The ramification is manifold, ranging from terrestrial communication relays to shipping/delivery services and disaster monitoring. To realize these applications, it is crucial to reliably control UAVs, while coping with unforeseen and risky events, e.g., inter-UAV collisions, warranting communication across UAVs. In this respect, several recent works consider a central entity assisting inter-agent message exchanges, referred to as a \emph{communication helper} \cite{miozzo2019coordinated}, and aim to reduce the payload sizes of inter-agent communication messages \cite{chen2018communication,shiri2020remote,roth2006communicate}. However, these works postulate that inter-UAV channel distributions and/or UAV state dynamics are known, which makes them ill-suited under dynamic real-time control environments, for instance under random wind perturbations.

 \begin{figure}[t!]
    \centering
    \includegraphics[width=.9\linewidth]{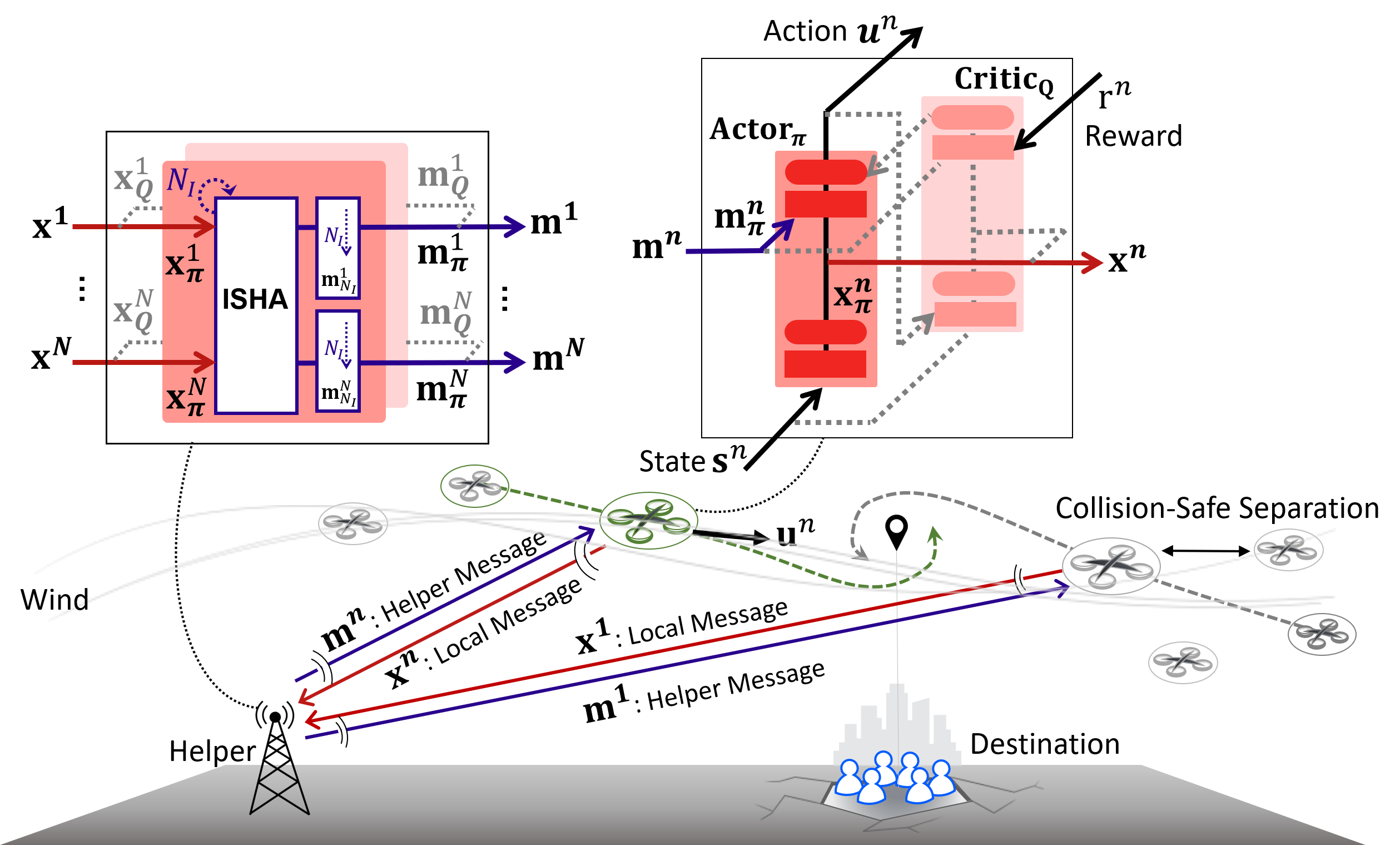}
    \caption{\small A helper-aided multi-UAV control system with an iterative single-head attention (ISHA) mechanism, where each UAV agent $n$ uploads its state embedding $ \mathbf{x}^n$, and the helper computes $\mathbf{m}^n$ using ISHA during $N_I$ iterations, which is downloaded by each agent.
} 
    \label{fig:system_model}
    \vspace{-14pt}
\end{figure}
  
Meanwhile, multi-agent deep reinforcement reinforcement learning (MADRL) has shown remarkable success in various real-time control applications~\cite{MADRL2020Wu}. By mapping each UAV into an agent, MADRL has a great potential in addressing the aforementioned real-time UAV control problems under dynamic environments. To operate MADRL, centralized training and distributed execution (CTDE) is a popular method, under which exchanging inter-agent communication messages (e.g., hidden-layer representations, local states, global reward, etc.) often yields significant control performance improvements \cite{foerster2016learning, OBL}. In this respect, one promising type of communication messages is the attention scores computed by the \emph{multi-head attention} (MHA) mechanism, which originally measure the inter-word relevance in a sentence, a paragraph, or a document \cite{vaswani2017attention}. Similarly, several recent works have applied attention to MADRL applications to measure the importance of inter-agent interactions \cite{iqbal2019actor,gu2021attention,Yang:IJCNN21}.

However, existing attention-based CTDE MADRL methods allow inter-agent communication only during the centralized training phase \cite{iqbal2019actor,gu2021attention,Yang:IJCNN21}, so can be brittle during the distributed control execution phase under dynamic environments. Furthermore, for multi-UAV path planning, these methods consider too many \cite{gu2021attention} or too few attention heads \cite{iqbal2019actor,Yang:IJCNN21}. By conducting numerical experiments, we found that a single attention head can attend to up to one other UAV agent (e.g., the nearest UAV), which may not be sufficient to avoid collision in a congested area (e.g., near the destination). On the other hand, with too many attention heads, the UAV may attend to spurious and weak dependencies (e.g., faraway UAVs), degrading control performance, not to mention the excessive computing cost.

To fill the aforementioned void, in this letter we propose a communication-aided MADRL framework with a novel \emph{iterative single-head attention} (ISHA) mechanism for real-time multi-UAV control. Under the actor-critic local model architecture of each agent, the ISHA located at a helper communicates not only with the critic models contributing only to the centralized training phase, but also with the actor models participating in distributed control execution. While agents upload their encoded state representations to the helper, ISHA iteratively calculates a set of attention scores over agents, which is downloaded by each agent for its local control decision-making. In contrast to MHA calculating attention scores by $N_H$ heads in parallel, ISHA sequentially produces attention scores with $N_I$ iterations using only a single head, in the hope of attending only to the top-$N_I$ most important neighboring agents while ignoring spurious attention scores. For a multi-UAV path planning scenario under random wind dynamics, simulation results corroborate that the proposed ISHA yields higher rewards with fewer MADRL training episodes, while achieving faster travel time and lower collision risks with smaller communication and computing energy footprints, compared to a baseline with MHA.

\section{System Model}\label{Sec:cent}

\subsection{Multi-UAV Control Model}
Consider a control system with a set $\mathcal{N}$ of $N$ UAVs, each of which is independently and randomly located inside a region $R$ at the same altitude $h$. Denote by $\mathbf{s}^{n}_{t} \in \mathcal{S}$ the UAV $n$'s state vector (e.g., location and velocity) at discrete time $t \geq 0$, and let $\mathbf{s}_t = (\mathbf{s}^1_t,\dots,\mathbf{s}^N_t) \in \mathcal{S}^N$, where $\mathcal{S}$ is the vector space of UAV states.
Moreover, denote by $\mathbf{u}^{n}_{t} \in \mathcal{U}$ the control action vector (e.g., acceleration magnitude and direction) of UAV $n \in \mathcal{N}$, where $\mathcal{U}$ is the vector space of control actions, and let $\mathbf{u}_t = (\mathbf{u}^1_t,\dots,\mathbf{u}^N_t) \in \mathcal{U}^{N}$. The goal of the system is to control each UAV to reach a common destination point inside $R$ while avoiding collisions as illustrated in Fig. \ref{fig:system_model}. 

\vspace{-10pt}
\subsection{Communication Network Model}
Each UAV sends its state information periodically to a \emph{communication helper} over an uplink (UL) channel and receive aggregated state information of the UAVs from the helper over a downlink (DL) channel. Suppose that each UAV-helper communication pair utilizes orthogonal frequency band to avoid inter-UAV interference and assume time division duplex (TDD) UL-DL configuration. 
The data rate of UAV-helper communication channel, at time $t$, is $R_t = {B \log_2 (1 + \frac{ P_{T}  }{P_{N} B}10^{-\frac{L_t}{10}})}$, where $B$ is the frequency bandwidth, $P_T$ is the transmission power, $P_N$ is the noise spectral density, and $L_t$ is the path-loss at time $t$ following the 3GPP urban-micro UAV channel model \cite{3GPP}. A single control interval is the time between the current and last control, assumed to be determined by the UL and DL communication delays while ignoring computing delays. We suppose that there is an upper limit of each UL or DL communication delay. If the delay exceeds the limit, the receiver (i.e., helper or UAV) utilizes its latest received information from the corresponding transmitter (i.e., UAV or helper).

\vspace{-10pt}

\subsection{MADRL Formulation}
Under the aforementioned communication and control scenario, we aim to optimize the control policy for each UAV so as to find a path with a low collision risk for by solving an MADRL problem. To begin with, we first formulate a baseline MADRL problem where each UAV agent receives state information from all the other UAVs in the region through the communication helper. Define by $\mathsf{P}^n:\mathcal{S}\times\mathcal{S}\times\mathcal{U} \rightarrow [0,1]$ the state transition function at agent $n \in \mathcal{N}$, and $\mathsf{r}_{t}^{n}: \mathcal{S}^N \times \mathcal{U}^N\to \mathbb{R}$ the reward at time $t$. Define by ${\pi}^n \!: \! \mathcal{S}^{N} \! \times \! \mathcal{U}\to [0, 1] $ the policy function (i.e., \emph{actor}) of   agent $ n $, where  $\pi^{n}(\mathbf{s}_t,\mathbf{u}_t^n)$ is the conditional probability of taking action $\mathbf{u}_{t}^{n}$ for given states $\mathbf{s}_{t}$. The objective of the MADRL problem is to learn policy models that maximize the agents' expected long-term reward. The reward $\mathsf{r}_{t}^{n}$ is a function of all agents' states $\mathbf{s}_t$ and actions $\mathbf{u}_t$, which is given by the sum of every positive reward $\mathsf{r}_{+,t}^{n}(\mathbf{s}_t^n) = \min \left\{  \frac{1}{\|\mathbf{s}_t^n - \mathbf{s}^g\|^{0.5}} , \mathsf{r}_{g}   \right\}$ and every negative reward $\mathsf{r}_{-,t}^{n}(\mathbf{s}_t^n) =  - \frac{1}{N}\sum_{m \neq n}\frac{1}{  \|\mathbf{s}_t^n - \mathbf{s}_t^m\|^{0.5}}$ that encourages the agents to reach the destination and discourages inter-UAV collisions, respectively. The term $\mathbf{s}^g$ is the target state after reaching the destination in a single agent scenario, $\mathsf{r}_{g}>0$ is a constant, and the initial reward $\mathsf{r}^n_0 = 0$, $\forall n\in\mathcal{N}$.

To enhance the performance of training by intentionally exposing   agents to   collision events, we leverage the soft actor-critic (SAC) algorithm \cite{haarnoja2018soft} which maximizes the trade-off between expected return and the randomness of the policy, i.e., entropy. Then, the corresponding sum expected long-term reward of all agents with fixed initial state $\mathbf{s} \in \mathcal{S}^N $ is
\begin{align}
{\mathsf{V}}^{\pi}(\mathbf{s})  = \sum_{n \in \mathcal{N}} \mathbb{E}\!\left[ \sum_{t=0}^{\infty} \gamma^{t} ( {\mathsf{r}}^{n}_{t}  +  \lambda \mathsf{h}^{\pi^n}_t )\!\mid\!\mathbf{s}_0 = \mathbf{s}\right],
\end{align}
where the expectation is taken with respect to the policy functions and state transition functions, $ \gamma \in ( 0, 1 ]  $ is a discount factor, $ \lambda \geq 0$ is the exploration and exploitation regularization coefficient,  and $\pi:\mathcal{S}^{N}\times\mathcal{U}^{N} \rightarrow [0,1]$ denotes a joint policy function of all agents, such that $\pi (\mathbf{s}_t, \mathbf{u}_t) = \prod_{n \in \mathcal{N}} {\pi}^{n} (\mathbf{s}_{t}, \mathbf{u}^{n}_{t})$ holds, $\forall \mathbf{s}_t \in\mathcal{S}^{N}$ and $\forall \mathbf{u}_t \in \mathcal{U}^N$, at time $t \!\geq \! 0$,
the function $\mathsf{h}_{\pi^n}(\mathbf{s}_t)$ regularizes the reward, which is defined with an information-theoretic entropy of the policy function with states $\mathbf{s}_t$, i.e., $\mathsf{h}^{\pi^n}_t \!=\! -\sum_{\mathbf{u}_t^{n}\in\mathcal{U}}\pi^{n}(\mathbf{s}_t,\mathbf{u}_t^{n})\log\pi^{n}(\mathbf{s}_t,\mathbf{u}_t^{n})$. 
{The training of SAC is realized by concurrent updating of policy and two quality functions.}
The quality function with fixed initial actions $\mathbf{u} \in \mathcal{U}^N$ and states $\mathbf{s} \in \mathcal{S}^N$ is $\mathsf{Q}^{\pi}\!(\mathbf{s},\!\mathbf{u}) \! = \!\sum_{n \in \mathcal{N}} \mathsf{Q}^{{\pi}^n}(\mathbf{s},\mathbf{u}^{n})$, where $\mathsf{Q}^{{\pi}^n}(\mathbf{s},\mathbf{u}^n)$ is the local quality function, i.e., \emph{critic}, at each agent $n$ which is defined as
\begin{align} \label{Eq:Rew_approx}
\mathsf{Q}^{{\pi}^n}(\mathbf{s},\mathbf{u}^n) \!=\!  \mathbb{E} \! \left[ \sum_{t=0}^{\infty} \gamma^{t} (   \mathsf{r}^{n}_{t} \!+\! \lambda \mathsf{h}^{{\pi}^{n}}_t)\!\mid\!\mathbf{s}_0 \!=\! \mathbf{s},\mathbf{u}_0^n \!=\! \mathbf{u}^n \right].
\end{align}
Then the expected quality over all initializations is 
\begin{align}\label{Eq:rew_cent}
\mathsf{J}^{\pi}  =  \sum_{\mathbf{s} \in \mathcal{S}^{N}} \mathsf{D}^{\pi}(\mathbf{s})  \sum_{\mathbf{u} \in \mathcal{U}^{N}} \pi(\mathbf{s},\mathbf{u}) \mathsf{Q}^{\pi} (\mathbf{s}, \mathbf{u}),
\end{align}
where $\mathsf{D}^{\pi}(\mathbf{s}) = \sum_{t=0}^{\infty} \gamma^{t} \bar{\mathsf{P}}^{\pi}(\mathbf{s}_{t}, \mathbf{s})$, and $\bar{\mathsf{P}}^{\pi}(\mathbf{s}_{t}, \mathbf{s})$ is the state transition function to $\mathbf{s}_{t}$ from the initial state $\mathbf{s}_0 = \mathbf{s}$ based on the joint policy function $\pi$. Now by maximizing \eqref{Eq:rew_cent}, we can obtain an optimal policy $\pi^*$.

{
The actor and critic of each agent $n$, requires the information about other agents, and the communication helper is considered to provide the agents with this information as shown in Fig. \ref{fig:system_model}.
We consider each agent $n$ has a pair of non-linear neural networks (NN) corresponding to its actor and critic NNs that encodes agent's state $\mathbf{s}_t^{n}$ and state-action pair $(\mathbf{s}_t^{n}, \mathbf{u}_t^{n})$ into representations (local messages) $\mathbf{x}^{n}_{\pi ,t}, \mathbf{x}^{n}_{Q ,t} \in \mathcal{X}$,
corresponding to the actor and critic NNs, where $\mathcal{X}$ is the space of representations. 
The helper gathers the local messages $\mathbf{x}^{n}_{t} \in \{  \mathbf{x}^{n}_{\pi,t}, \mathbf{x}^{n}_{Q,t}  \} $ and sends back
the corresponding representations set $  \{ \mathbf{x}^{m}_{t}\!\!:\!\! m \!\in\! \mathcal{N} ~ \text{and} ~ m \!\neq\! n \}$ to each agent $n$ through $-\log_{2}|\mathcal{X}|^{N-1}$ bits of information at the DL. Then, each agent $n$, after receiving the messages from the helper, concatenates them with its local message $\mathbf{x}^{n}_{t}$ and feeds them to the corresponding actor NN which approximates $\pi^{n}(\mathbf{s}_t,\mathbf{u}_t^{n})$, and critic NN which approximates  $\mathsf{Q}^{{\pi}^n}(\mathbf{s},\mathbf{u}^{n}) $, in order to output action $\mathbf{u}_t^{n}$ and critic value $ \mathsf{Q}^{\pi^n}$, respectively. 
The Vanilla method incurs a large communication cost and lacks scalability against the increasing number of agents $N$.
}

\section{Attention mechanism for Obtaining Meaningful Partial State Representation}
The communication payload and scalability issues in the {Vanilla MADRL} can be improved by processing the state information received at the helper using a well-designed message generating vector function $\mathbf{m}^n \! : \! \mathcal{X}^N \!\! \to \!\! \mathcal{M}$, and transmitting the messages $ \mathbf{m}^{n}(\mathbf{x}_{\pi,t})$ and $\mathbf{m}^{n}(\mathbf{x}_{Q,t})$ for agent $n$'s actor and critic NNs, respectively, where $\mathbf{x}_{\pi,t} \!=\! \{ \mathbf{x}_{\pi,t}^{n} \!:\! n \in \mathcal{N} \}$, $\mathbf{x}_{Q,t} \!=\! \{ \mathbf{x}_{Q,t}^{n} \!:\! n \in \mathcal{N} \}$, and $\mathcal{M}$ denotes the message space. For simplicity, we use $\mathbf{m}^n_{\pi,t}$ and $\mathbf{m}^n_{Q,t}$ instead of $ \mathbf{m}^{n}(\mathbf{x}_{\pi,t})$ and $\mathbf{m}^{n}(\mathbf{x}_{Q,t})$, respectively, hereafter. Note that this approach becomes communication-efficient if each helper message $ \mathbf{m}^n_{t} \in \{ \mathbf{m}^n_{\pi,t}, \mathbf{m}^n_{Q,t} \} $ produced by the message function represents the states of the agents partially, i.e., $|\mathcal{M}| \!\leq\! |\mathcal{X}|^{N-1}$. Now, consider a new policy function $\hat{\pi}^{n}: \mathcal{S} \times \mathcal{M} \times \mathcal{U} \to [0, 1]$ at the agent $n\in\mathcal{N}$, which takes the partial state representation received from the helper as an input instead of the full state representation as in the Vanilla baseline. Then, by maximizing the expected quality $\mathsf{J}^{\hat{\pi}}$, we obtain a near-optimal policy $\hat{\pi}^*$ which approximates the optimal policy $\pi^*$. Note that the optimality (effectiveness) of $\hat{\pi}^*$ trades off the communication-efficiency of this scheme and is bounded by the optimal policy of baseline MADRL $\pi^*$.

\vspace{0pt}

\subsection{Baseline - Multi-Head Attention}
\begin{figure}[t]
\centering
\subfigure{\includegraphics[width=0.48\textwidth, trim={0cm 0cm 0.5cm 0cm},clip]{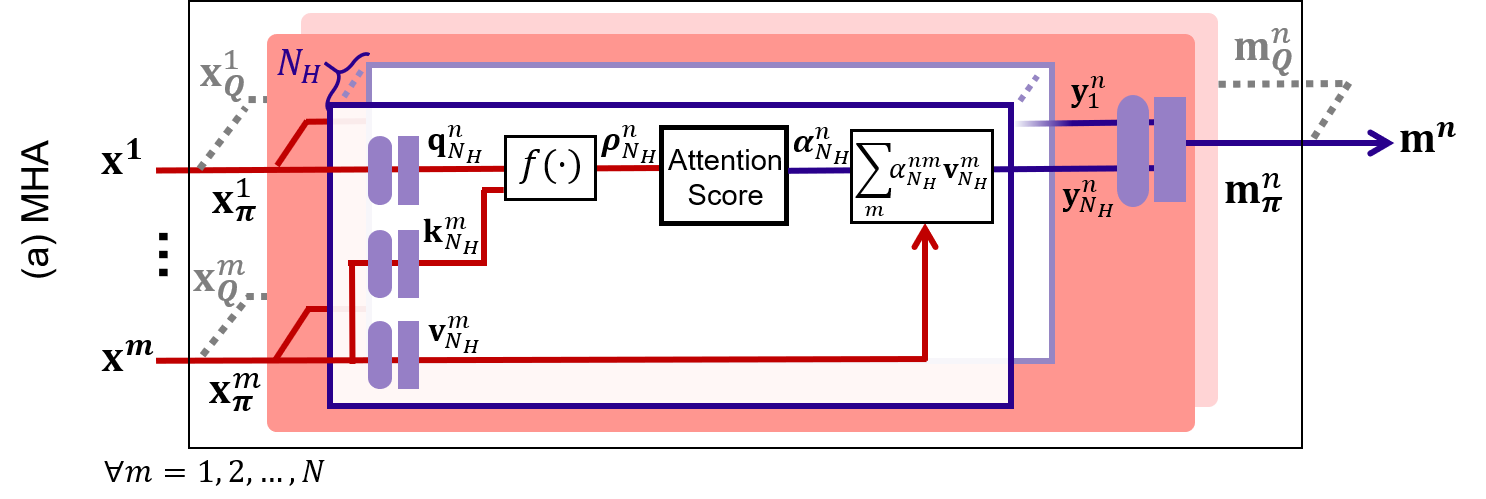}\label{fig:MHAtt}}
\subfigure{\includegraphics[width=0.48\textwidth, trim={0cm 0cm 0.5cm 0cm},clip]{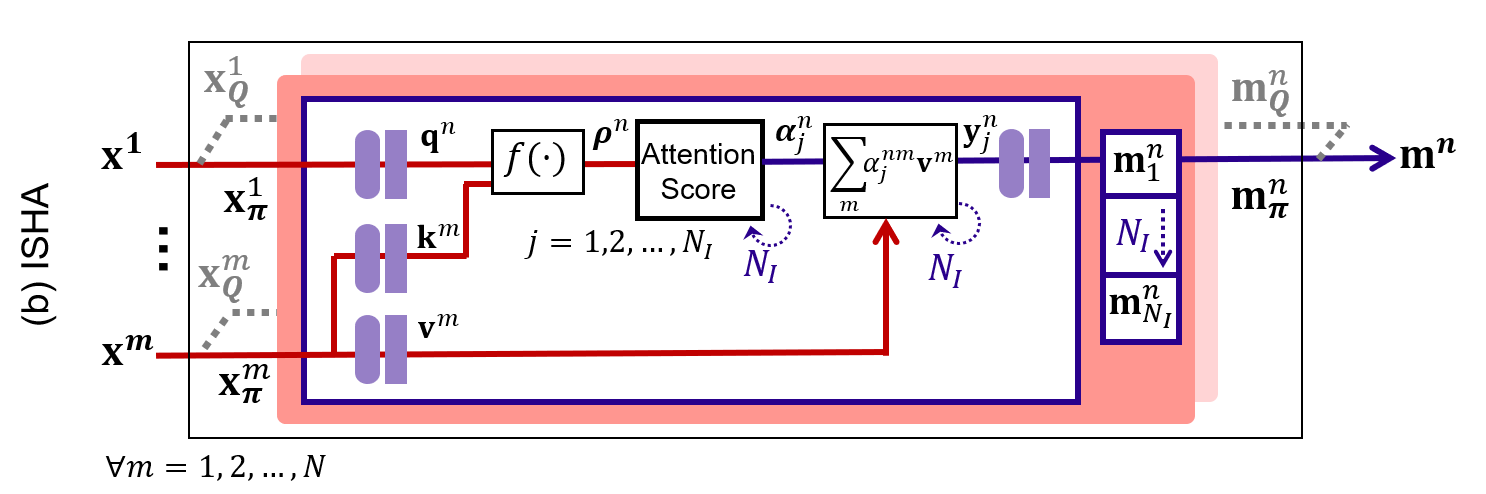}\label{fig:ISHAtt}}
\vspace{-5pt}
\caption{\small The operations of the proposed (b) ISHA mechanism with $N_I$ iterations, compared with the (a) MHA mechanism with $N_H$ heads.}
\label{fig:IAtt}
\vspace{0pt}
\end{figure}

One promising way of designing such a message generation function is leveraging the MHA structure \cite{vaswani2017attention}. Fig. \ref{fig:MHAtt}, illustrates the simplified MHA structure obtaining a message $\mathbf{m}^{n}_t$ for the agent $n$ by concatenating and feeding $N_H$ sub-messages $\mathbf{y}^{n}_{c,t}$, $\forall c  \!\in\! \{1,\dots,N_H\}$ to an output NN. The sub-messages are calculated in $N_H$ parallel single-head attention structure as a weighted sum $\mathbf{y}_{c, t}^{n} = \sum_{m \neq n} {\alpha}_{c,t}^{n,m} \mathbf{v}^{m}_{c,t} $, where the \emph{values} $\mathbf{v}^{m}_{c,t}$ are obtained by feeding the state representation $\mathbf{x}^{m}_{t}$ into a linear NN followed by a normalization layer $\forall m\in \mathcal{N}$, and the \emph{attention scores} ${\alpha}_{c,t}^{n,m}$ for $c\!\in\!\{1,\dots,N_H\}$ are calculated as follows. The \emph{query} $\mathbf{q}^{n}_{c,t}$ and \emph{key} $\mathbf{k}^{n}_{c,t} $ are obtained in a similar way to the values $\mathbf{v}^{n}_{c,t}$.
From the obtained queries and keys, the relevance score ${\rho}_{c,t}^{n,m}$ between agent $n$ and the other agents at time $t$ in each process $c \in \{1, \ldots, N_H \}$ is calculated using a relevance score function $f(\cdot)$ which takes a query and key as an input and outputs a scalar real value \cite{hu2019introductory}. Note that the relevance score measures the similarity between the pairs $(\mathbf{q}^{n}_{c,t}, \mathbf{k}^{m}_{c,t})$, that is the interaction level between UAV $n$ and UAV $m$. In this paper,
the relevance score is obtained using the dot product function.
Then, the attention scores ${\alpha}_{c,t}^{n,m} $
are obtained using the softmax function. Overall, $N$ parallel multi-head attentions run concurrently to obtain $N$ messages $\mathbf{m}_t^1, \ldots, \mathbf{m}_t^N$ for $N$ agents respectively.
For both actor and critic of agent $n$, the corresponding received message $ \mathbf{m}^{n}_{t}$ from the helper is concatenated with its local message $ \mathbf{x}^{n}_{t}$ and fed into the corresponding NNs that outputs the policy ${\pi^n}$ and quality $ \mathsf{Q}^{\pi^n}$ as illustrated in Fig. \ref{fig:system_model}.

\begin{figure}[t]
\centering
\subfigure{\includegraphics[width=0.24\textwidth, trim={0cm 0cm 0.5cm 0cm},clip]{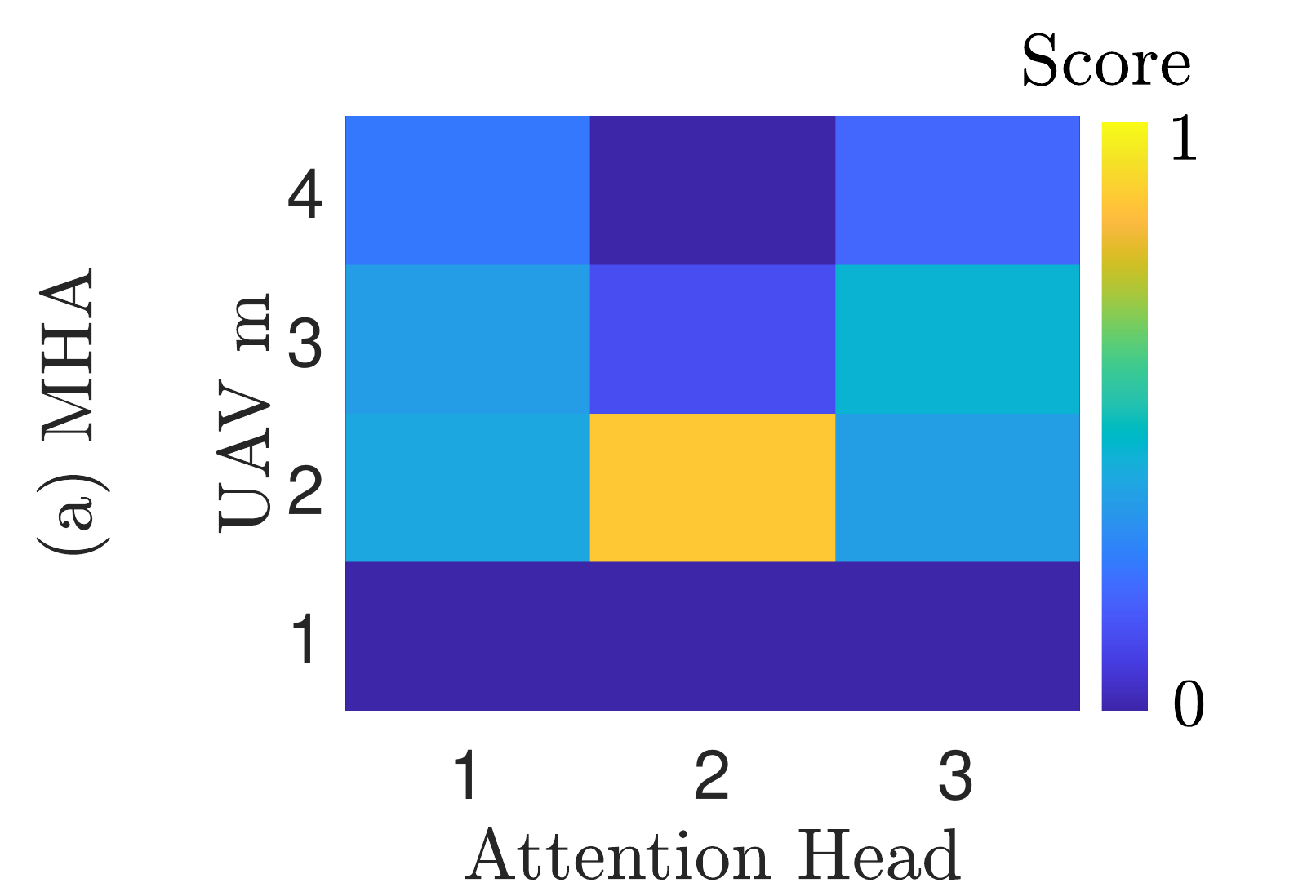}\label{fig_heatmapMHA}}
\subfigure{\includegraphics[width=0.24\textwidth, trim={0.5cm 0cm 0cm 0cm},clip]{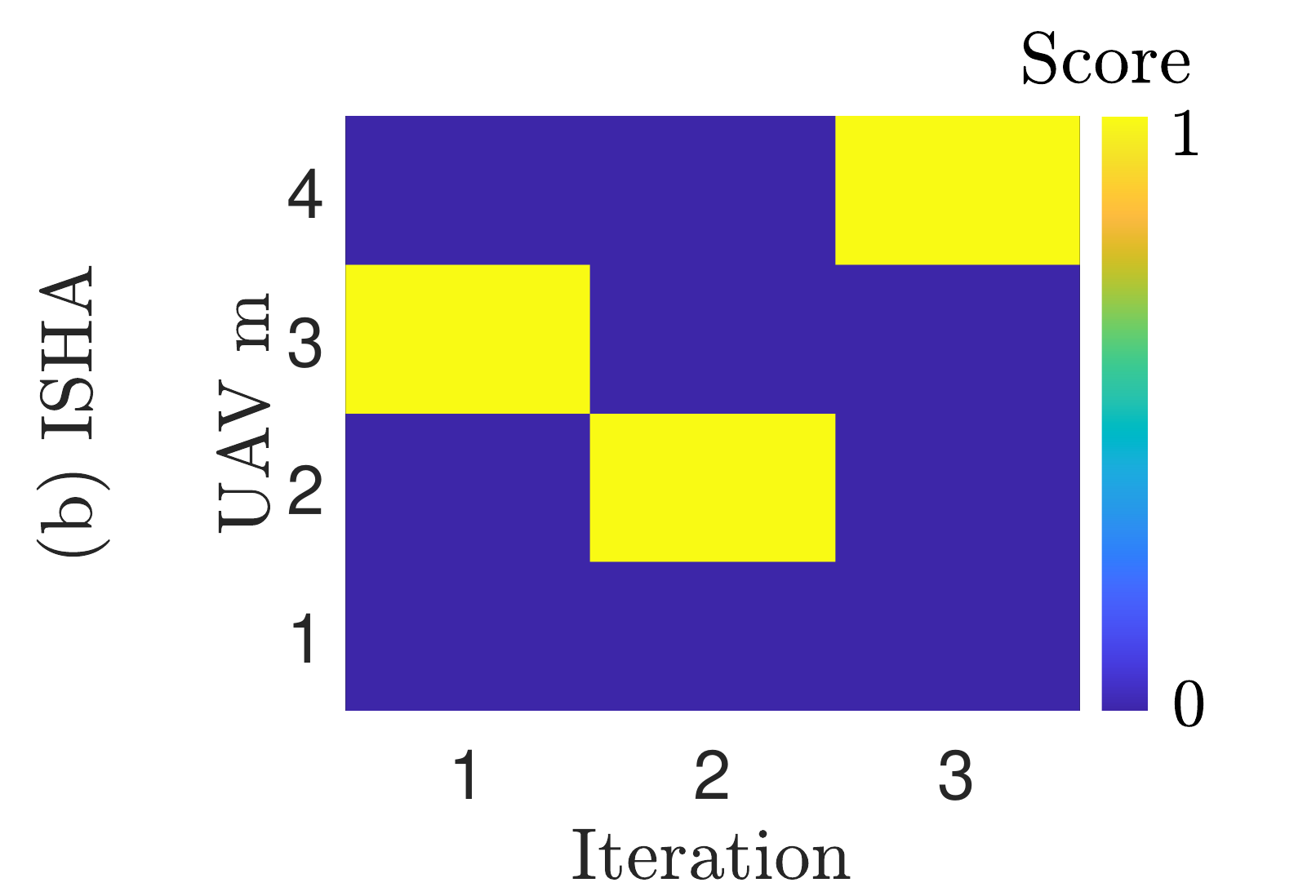}\label{fig_heatmapISHA}}
\caption{Attention scores obtained for an agent by (a) MHA, and (b) ISHA methods, in a scenario with $N=4$ agents.}
\label{fig:Att_Scores}
\vspace{-12pt}
\end{figure}

\subsection{Proposed - Iterative Single-Head Attention}
The MHA-based method with $N_H$ heads is proper for discovering the long-range dependencies, and it can attend to state representations of $N_H$ different agents in maximum. Then is not efficient for multi-UAV control scenarios with short-range inter-UAV dependencies and limited communication resources.
Here, we propose the ISHA structure that can attend to as many messages as the number of iterations with less processing cost than MHA, obtaining short-range dependencies at each message.
For example, Fig. \ref{fig_heatmapMHA} shows a heatmap of  attention scores obtained at each attention head of random agent using the MHA structure with $N_H = 3$ parallel heads in a scenario with $N=4$ UAVs. Here, UAV $2$'s state representation is attended by the attention head $2$, and the UAV $3$'s state representations get high attention scores from both heads $1$ and $3$, but UAV $4$'s state representation is not being attended well. However, the ISHA method can obtain more distinct attention scores as shown in Fig. \ref{fig:Att_Scores}(b) ordered by the inter-UAV short-range dependencies.

Refer to Fig. \ref{fig:ISHAtt} which illustrates the structure of ISHA. In ISHA, similar to the process of MHA, the gathered state representations ${\mathbf{x}}^{n}_{t}$ from all agents are used to obtain the query $ {\mathbf{q}}^{n}_{t} $, key $ {\mathbf{k}}^{n}_{t} $, and value $ {\mathbf{v}}^{n}_{t} $, by feeding them into the linear NNs followed by a normalization layer and the dot function is used to calculate the relevance scores $ {\boldsymbol{\rho}}_{t}^{n} = \{ {\rho}_{t}^{n,1}, \ldots, {\rho}_{t}^{n,N} \}  $ for each agent $n$. 
{
The relevance scores ${\boldsymbol{\rho}}_{t}^{n}$ for each agent $n$ captures the inter-UAV interactions, and can be utilized to attend to several messages if there is enough communication resources. To do so, we obtain $N_I$ attention score sets  $\boldsymbol{\alpha}_{c,t}^n = \{ {\alpha}_{c,t}^{n,m} : m = 1, \ldots , N \} $, for $c = 1, \ldots, N_I $, by an elimination process shown in Algorithm 1, and then obtain $N_I$ corresponding messages by a weighted sum $\mathbf{y}_{c, t}^{n} = \sum_{m} {\alpha}_{c,t}^{n,m}  {\mathbf{v}}^{m}_{t} $, $\forall c\in\{1,\dots,N_I\}$ . 
In this algorithm, the set $\mathcal{E}^{n}$ is defined to store the eliminated elements, and it is initialized as $\mathcal{E}^{n}=\{n\}$ to remove the self-attention for the corresponding agent $n$. 
At each iteration $c$ of the algorithm, one attention score set $\boldsymbol{\alpha}_{c,t}^n $ is calculated as
\begin{equation}
{\alpha}_{c,t}^{n,m}   = 
    \begin{cases}
          \exp{\left(\frac{{\rho}_{t}^{n,m}}{\beta}   \right)}/ \sum_{j \not\in \mathcal{E}^{n} } \exp{ \left(\frac{{\rho}_{t}^{n,j}}{\beta} \right) },    & {m \not\in \mathcal{E}^{n}}
        \\
         0, & {m \in \mathcal{E}^{n}}    
    \end{cases} 
    \label{Eq:alpha}
\end{equation}
where the temperature coefficient $\beta$ is a hyper-parameter used to tune the normalization in order to prevent very small gradients and help the training of the algorithm.
}

Therefore, by this process, we obtain $N_I$ weighted sums $\mathbf{y}_{c, t}^{n}$, $\forall c\in\{1,\dots,N_I\}$ which are differentiable and  ordered by the importance of messages. 
Then, each weighted sum $\mathbf{y}_{c, t}^{n}$ is fed into a non-linear NN with a smaller output layer size than input layer size, to obtain the message $\mathbf{m}_{c,t}^{n}$ which has a smaller size than $\mathbf{y}_{c, t}^{n}$, in order to reduce the communication payload size.  
Finally the output semantic message $ \mathbf{m}^{n}_{t} $ is defined as a $N_I$-tuple consisting of the messages $ \mathbf{m}_{c,t}^{n}$, i.e., $ \mathbf{m}^{n}_{t} = (\mathbf{m}_{1,t}^{n}, \ldots, \mathbf{m}_{N_I,t}^{n}) $.  In other words, the semantic message includes $N_I$ sub-messages, where each of them attends to an interaction. Moreover, in both actor and critic of agent $n$, the semantic message $ \mathbf{m}^{n}_{t} $ is concatenated with the corresponding local message ${\mathbf{x}}^{n}_{t}$ to output the policy  ${\pi^n}$ and quality $ \mathsf{Q}^{\pi^n}$ as denoted in Fig. \ref{fig:system_model}.

\begin{algorithm}[t]
	\caption{Iterative Attention Score Calculation}
	\begin{algorithmic}[1]
		\STATE \textbf{Initialization:} \textbf{Read} $ {\boldsymbol{\rho}}_{t}^{n}  $;  \textbf{set} $\mathcal{E}^{n} = \{n\}$, and $c = 1$.
		
		\WHILE {$  c \leq  \nu $}

		\STATE {\textbf{Calculate} and \textbf{Output} score set $\boldsymbol{\alpha}_{t,c}^{n}$ using Eq. \ref{Eq:alpha}}.
		
		\STATE \textbf{Eliminate} maximum element $m^{*} \!=\! \argmaxA_{m} \boldsymbol{\alpha}_{t,c}^{n}$ by $ \mathcal{E}^{n} \! \leftarrow \! \mathcal{E}^{n}  \cup \{ m^{*} \} $.

		\STATE $ c \leftarrow c + 1 $.

		\ENDWHILE
		
	\end{algorithmic} \label{alg:01}
\end{algorithm}

\section{Simulations}
In this section, we investigate the training, control, and communication performance of the proposed ISHA method with $N_I=1$ (ISHA$_1$) and $N_I = 3$ (ISHA$_3$), in comparison with MHA with $N_H = 3$ (MHA$_3$) and the Vanilla baselines, by simulating the control of $ N=4 $ UAVs in an area of $ 1000 \times 1000 ~ \text{m}^2 $. The ground base-station (BS) as the helper is located at a fixed location $(1000, 1000) ~ \text{m}$, and the flying height of agents is fixed at $50$ \text{m}. Two multi-layer perceptrons (MLP) with 3 hidden layers of sizes $100$ are used as the actor and critic, and two MLP with 2 hidden layers of sizes $63$ and one output layer of size $63$ are used as the local message generation functions of corresponding actor and critic at each agent, all with ReLU \cite{agarap2018deep} activation function. 
At the helper, linear NNs of size $ 63 \times 63$ are defined as query, key, and value NNs for all methods. The output NN for ISHA and MHA are defined by linear NNs of sizes  $ 63 \times 21$ and $63 \times 63$, respectively. The input layer size of the actor or critic NN at each agent is the sum of the local message and helper message dimensions. Other parameters are set as follows: carrier frequency is $2 \text{GHz}$; $B = 5 \text{MHz}$; $ P_N = - 100  \text{dBm}/\text{Hz} $; UL message size is $  63 \! \times \! 32~\text{bits}$;  DL message size are $ 21  \!\times\! 32~\text{bits}$, $  3 \times 21 \times 32~\text{bits}$,  $  63 \times 32~\text{bits}$, $  3 \times 63  \!\times\! 32~\text{bits}$ for ISHA$_1$, ISHA$_3$, MHA$_3$, and Vanilla respectively; $\mathbf{s}_g = \boldsymbol{0}$; and  $\mathsf{r}_{g} = 1$.

\begin{figure}[t]
\centering
\subfigure{\includegraphics[width=.8\columnwidth, trim={0cm 1.3cm 0cm 0cm},clip]{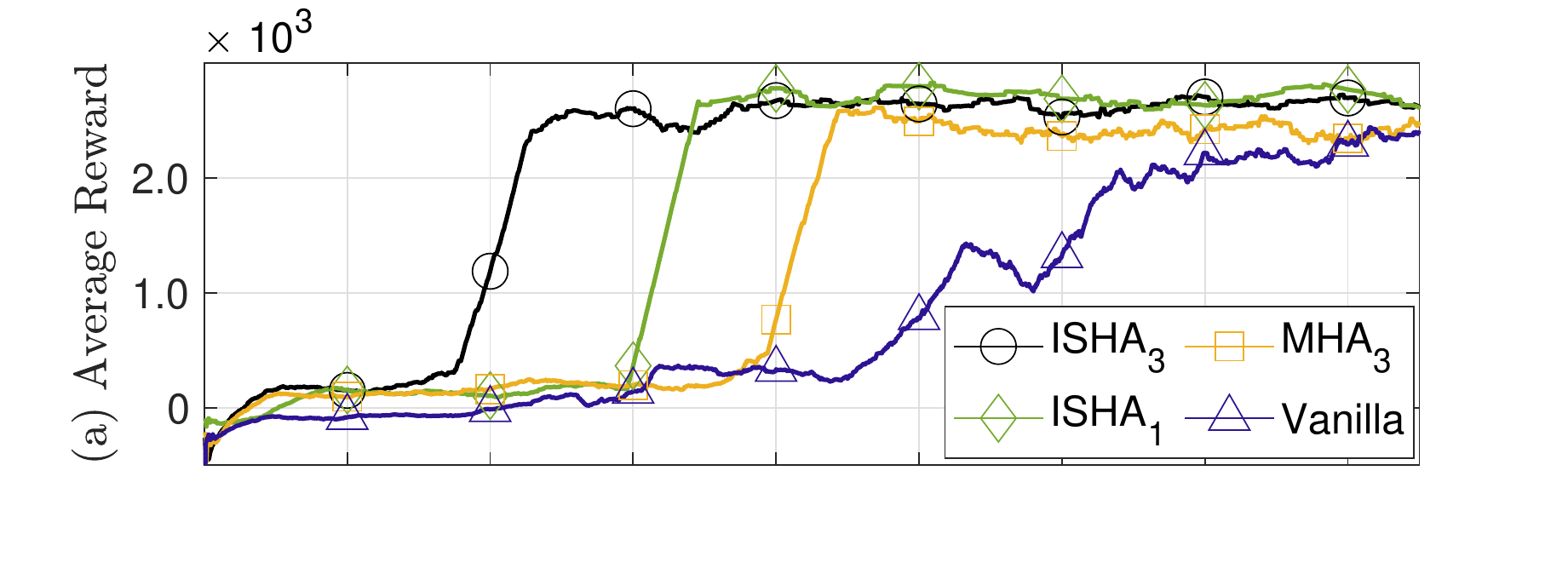}}
\subfigure{\includegraphics[width=.8\columnwidth, trim={0cm 0cm 0cm 0cm},clip]{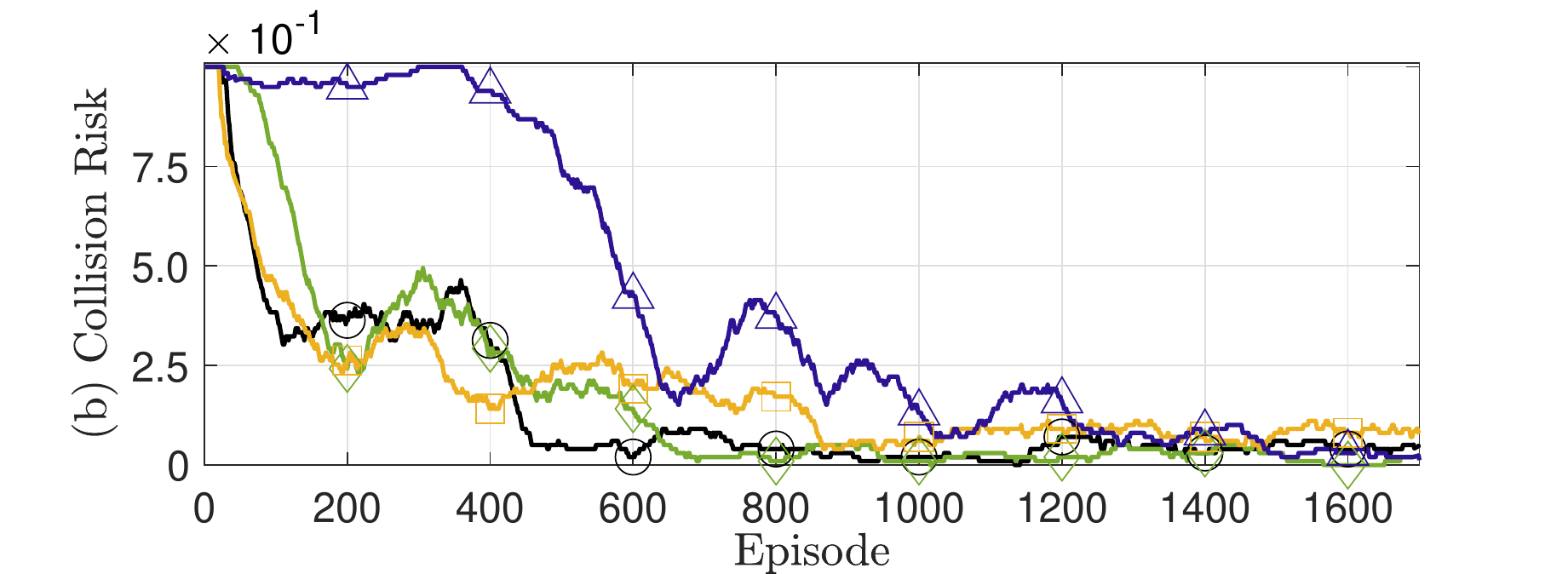}}
\caption{\small Comparison of average  reward and collision rate during training with ISHA$_3$, ISHA$_1$, MHA$_3$, and Vanilla.}
\label{fig:Training}
\vspace{-10pt}
\end{figure}

\begin{table}[t]
\centering
\caption{\small Model size and communication cost (bits per round).} 
\label{tab:CompComm}
\begin{tabular}{lllll}
\toprule
\multirow{2}{*}{\textbf{Methods}}   & \multicolumn{2}{c}{\textbf{Model Size}}  & \multicolumn{2}{c}{\textbf{Commun. Cost}}  \\
    &  Agent  &  Helper  & Agent  &  Helper    \\  \cmidrule{1-5} 
Vanilla   & $ 2.14 \!\times\! 10^5 $  & $0$ & $8.06 \!\times\! 10^3$ & $2.42 \!\times\! 10^4$ \\
MHA$_3$   & $ 1.63 \!\times\! 10^5$  & $1.58 \!\times\! 10^5$ & $8.06 \!\times\! 10^3$ & $8.06 \!\times\! 10^3$ \\
ISHA$_3$  & $ 1.63 \!\times\! 10^5 $ & $6.35 \!\times\! 10^4$ & $8.06 \!\times\! 10^3$ & $8.06 \!\times\! 10^3$ \\ 
ISHA$_1$  & $ \mathbf{1.47 \!\times\! 10^5} $ & $\mathbf{5.29 \!\times\! 10^4}$ & $\mathbf{8.06 \!\times\! 10^3}$ & $\mathbf{2.69 \!\times\! 10^3}$ \\ 

\bottomrule
\end{tabular}
\vspace{-10pt}

\end{table}

\begin{figure}[t]
\centering
\subfigure[ ISHA$_3$.]{\includegraphics[width=0.21\textwidth]{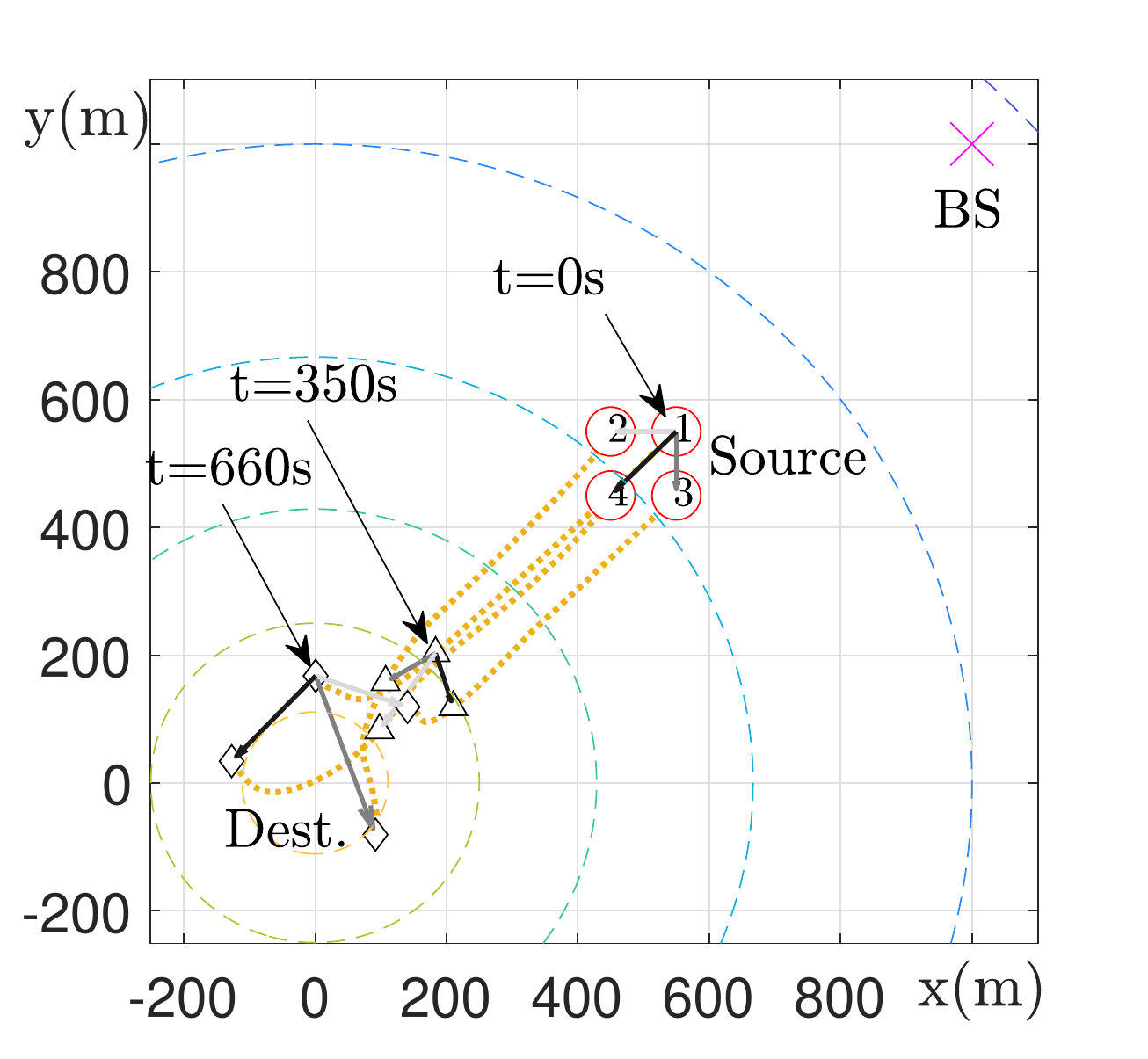}}
\subfigure[ISHA$_1$.]{\includegraphics[width=0.21\textwidth]{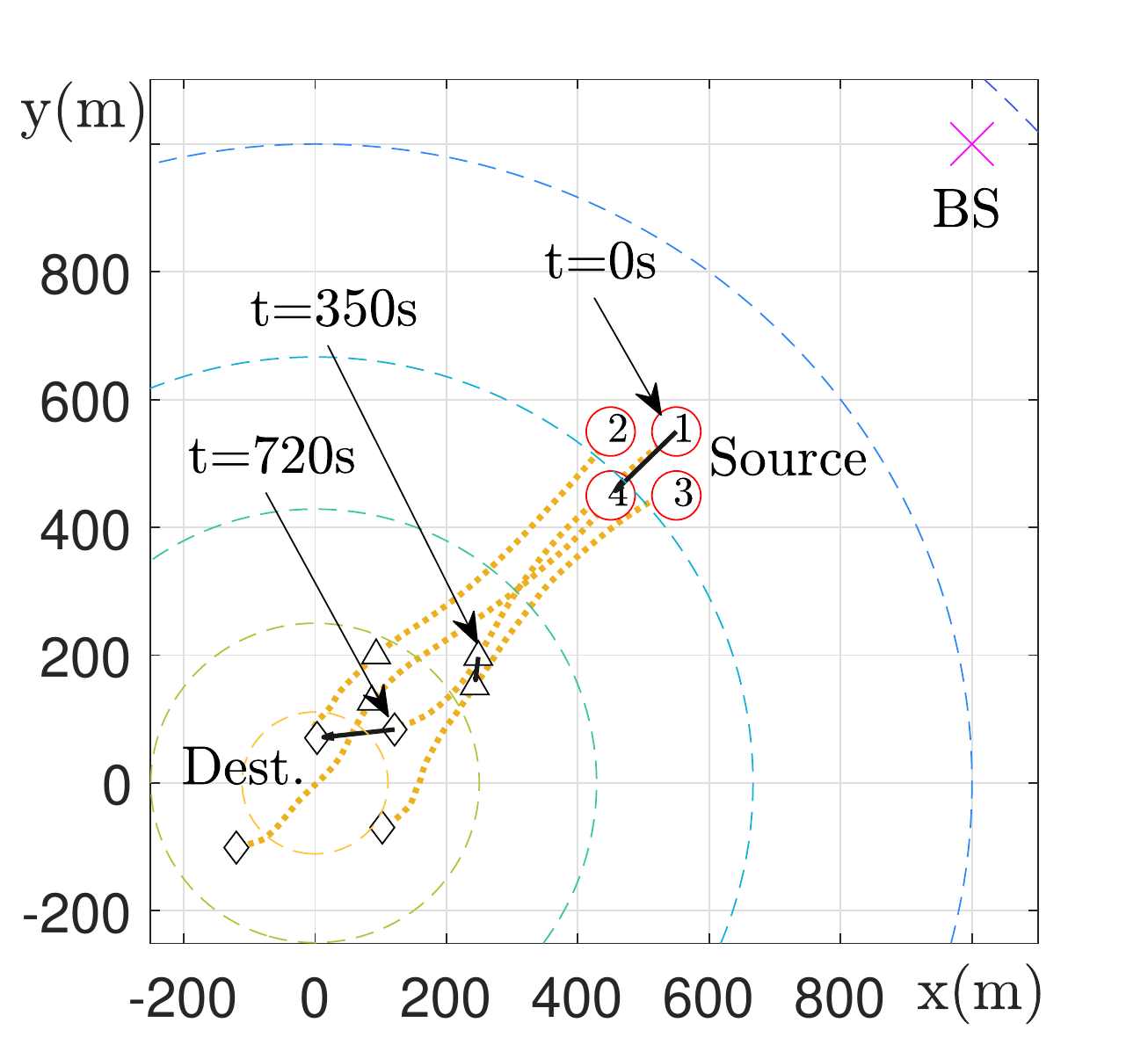}}
\subfigure[ MHA$_3$.]{\includegraphics[width=0.21\textwidth]{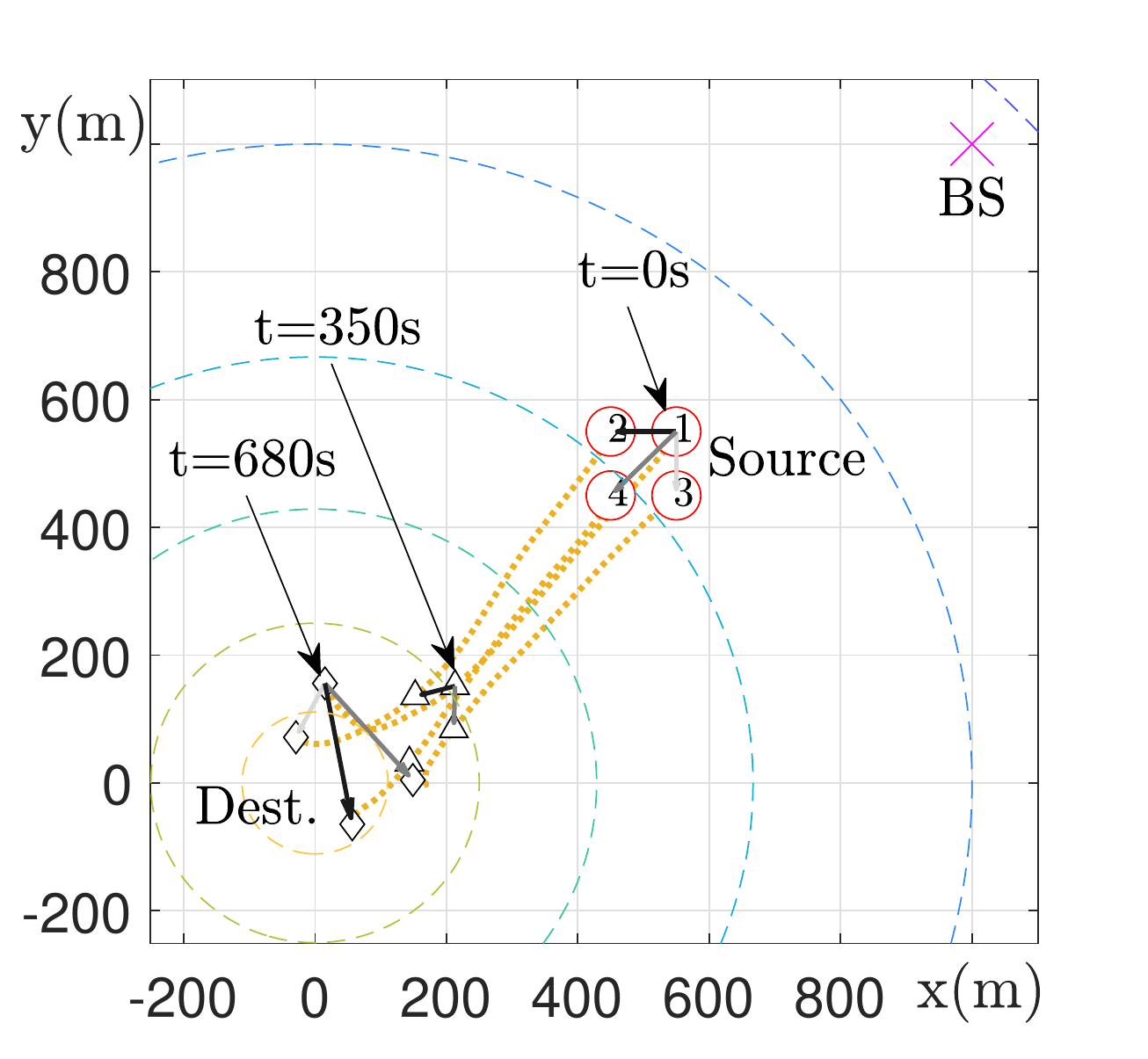}}
\subfigure[Vanilla]{\includegraphics[width=0.21\textwidth]{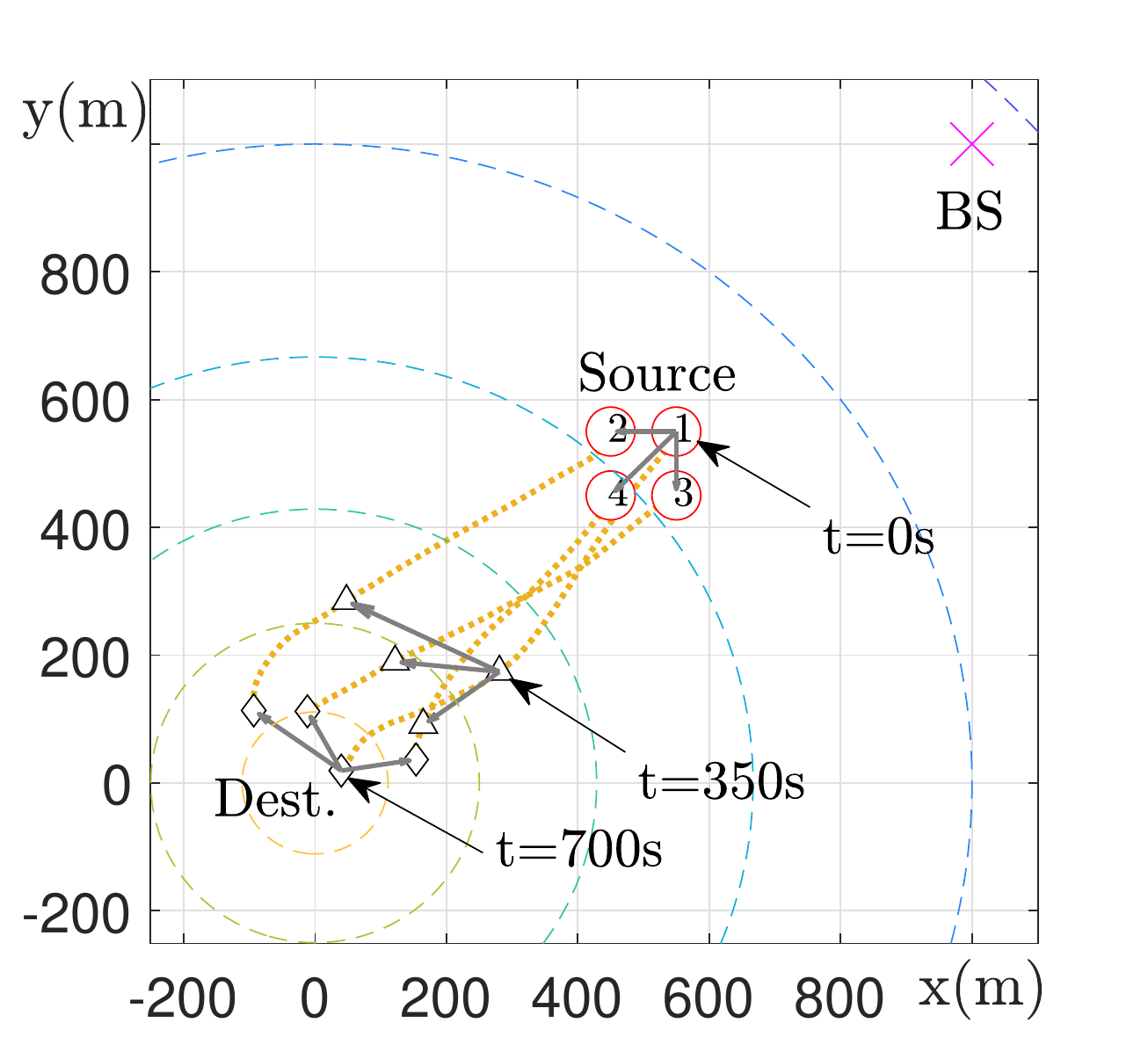}}
\caption{\small Trajectories for  different control methods.}
\label{fig:Trajectories}
\vspace{-10pt}
\end{figure}

\textbf{Training Time}.\quad
Fig. \ref{fig:Training} shows the average cumulative reward and collision risk of ISHA, MHA and Vanilla over training time, under perfect communication channel amd fixed control interval of $ 0.1 \text{s}$. The ISHA$_3$  method converges after about $ 500$ episodes, which is much faster than the other methods, since the helper can calculate meaningful information about the inter-UAV interactions with smaller model size compared to MHA and Vanilla (see Table \ref{tab:CompComm}). Furthermore, compared to ISHA$_3$,  ISHA$_1$ has smaller model sizes both at the helper and agents, but takes longer time for actors and critics to learn to extract inter-UAV information from the helper messages, even though it still converges faster than MHA and Vanilla.

\begin{figure}[t!]
\centering 
\includegraphics[width=0.85\columnwidth]{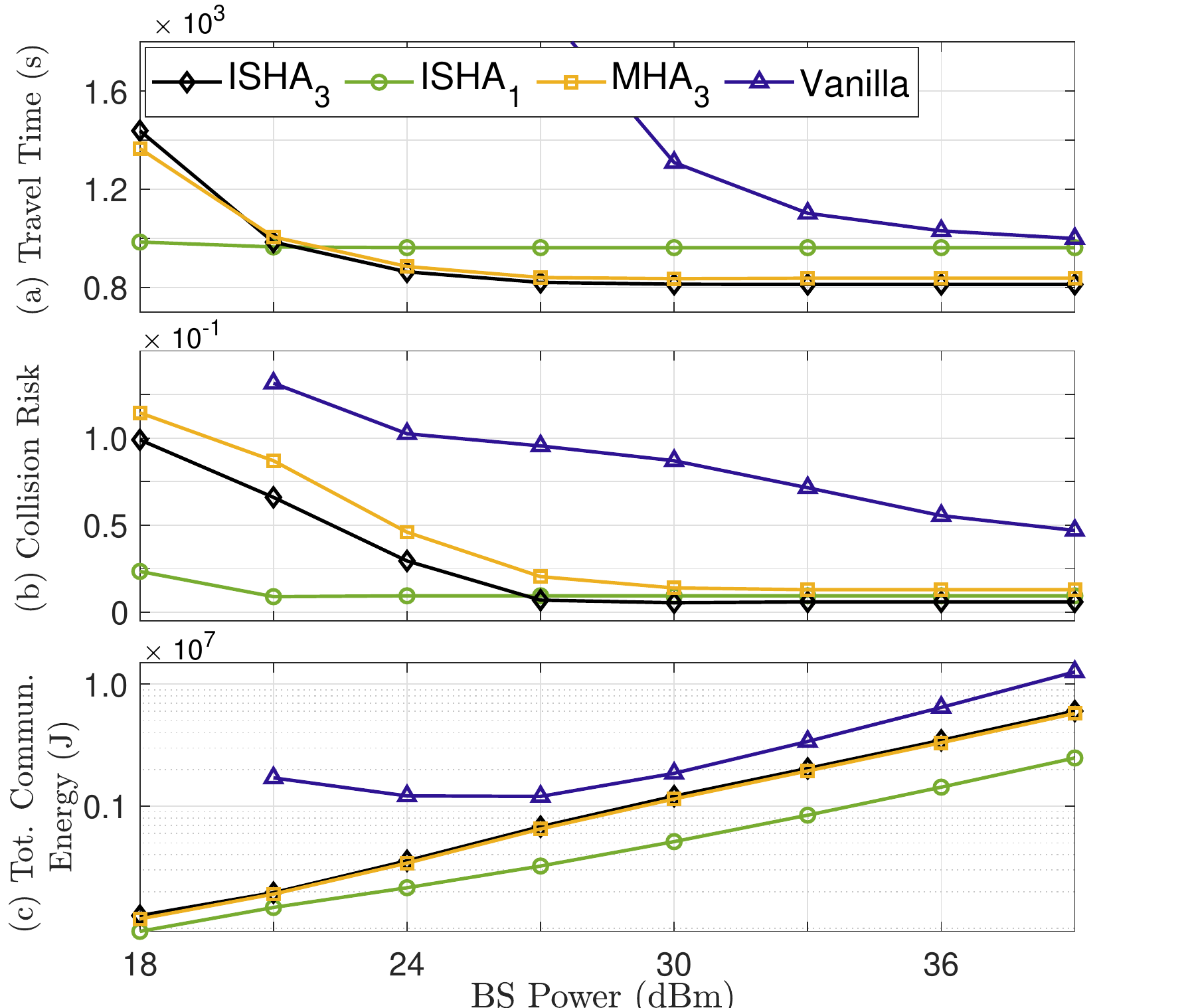}
\caption{\small Comparison of ISHA, MHA and Vanilla under varying helper transmission power.}
    \label{fig:Comparisons}
\vspace{-10pt}
\end{figure}

\textbf{Travel Time}.\quad
Fig. \ref{fig:Trajectories} represents the test trajectories of ISHA, MHA and Vanilla methods with UAVs initially located close to each other, under perfect communication channel with fixed control interval $ 0.1 \text{s}$. 
The effect of UAV's messages on the helper message calculated for a randomly selected UAV depicted using arrows; the darker arrow means more effect.
We  observe that ISHA$_3$ has the shortest travel time, and the reason is that, contrary to MHA$_3$, it always attends to $3$ distinct short-range interacting UAVs as shown in snapshot $t= 350 \text{s}$ in Fig. \ref{fig:Trajectories}(a) and Fig. \ref{fig:Trajectories}(c).  
The travel time performance of ISHA is better than the other baseline methods even with limited communications resources as shown in Fig. \ref{fig:Comparisons}(a), which illustrates the average travel time of different methods versus BS transmission powers under a maximum control interval of $1\text{s}$. 
Specifically, at  small helper transmission power, ISHA$_1$ can still maintain its travel time performance, while other methods cannot operate due to their higher communications requirements as specified by Table \ref{tab:CompComm}.

\textbf{Collision Risk}.\quad
On the other hand, ISHA methods achieve smaller collision risk both at the training as in Fig. \ref{fig:Training}(b) and execution in Fig. \ref{fig:Comparisons}(b). At BS transmission power higher than $27$dBm, ISHA$_3$ can achieve $2$ times smaller collision risk than its counterpart MHA, because of successful transmission of extracted short-range UAV interaction messages to the agents from the helper. However, at transmission powers less than $27$dBm, collision risk can be maintained below $0.025$ by utilizing ISHA$_1$ which requires smaller transmission bits (refer to Table. \ref{tab:CompComm}) for only 1 iteration of the algorithm.

\textbf{Communications Cost}.\quad
One remarkable observation from Fig. \ref{fig:Comparisons}(c) is that while the control performance of the baseline methods declines at low power settings, i.e., $ \leq 27 \text{dBm}$, the ISHA$_1$ can still obtain low collision risk path planning with smaller total communication energy consumption compare to other methods. The reason for this is two-fold, first the smaller communication message sizes as in  Table. \ref{tab:CompComm}, and second faster mission completion time as shown in Fig. \ref{fig:Comparisons}(a).

\section{Conclusion}
In this letter, we proposed an MADRL framework with a novel iterative single-head attention (ISHA) mechanism for communication-efficient multi-UAV path planning and control. Compared to the standard MHA which incurs spurious attention scores and excessive computing overhead, the proposed ISHA enables to capture important attention scores for control, thereby achieving faster mission completion while reducing total communication and computing costs.

\bibliographystyle{IEEEtran}
\bibliography{IEEEabrv,ref.bib}

\end{document}